# Brain MRI Segmentation using Rule-Based Hybrid Approach

Mustansar Fiaz
*School of Computer Science and Engineering*
*Kyungpook National University*
Daegu, South Korea
mustansar@vr.knu.ac.kr

Kamran Ali
*Department of computer science*
*University of Central Florida*
Orland, FL, USA
kamran@nights.ucf.edu

Abdul Rehman
*School of Computer Science and Engineering*
*Kyungpook National University*
Daegu, South Korea
a.rehman.iiui@gmail.com

M. Junaid Gul
*School of Computer Science and Engineering*
*Kyungpook National University*
Daegu, South Korea
junaidgul@live.com.pk

Soon Ki Jung*
*School of Computer Science and Engineering*
*Kyungpook National University*
Daegu, South Korea
skjung@knu.ac.kr

*Abstract*—Medical image segmentation being a substantial component of image processing plays a significant role to analyze gross anatomy, to locate an infirmity and to plan the surgical procedures. Segmentation of brain Magnetic Resonance Imaging (MRI) is of considerable importance for the accurate diagnosis. However, precise and accurate segmentation of brain MRI is a challenging task. Here, we present an efficient framework for segmentation of brain MR images. For this purpose, Gabor transform method is used to compute features of brain MRI. Then, these features are classified by using four different classifiers i.e., Incremental Supervised Neural Network (ISNN), K-Nearest Neighbor (KNN), Probabilistic Neural Network (PNN), and Support Vector Machine (SVM). Performance of these classifiers is investigated over different images of brain MRI and the variation in the performance of these classifiers is observed for different brain tissues. Thus, we proposed a rule-based hybrid approach to segment brain MRI. Experimental results show that the performance of these classifiers varies over each tissue MRI and the proposed rule-based hybrid approach exhibits better segmentation of brain MRI tissues.

## I. INTRODUCTION

In the last few decades, a significant contribution has been made by medical image processing in various medical technologies such as X-rays, Computerized Tomography (CT) scan, and Magnetic Resonance Imaging (MRI) etc. X-ray is the primitive technique used for medical investigation [1]. High energy and short wavelength electromagnetic waves are used in X-ray [2]. CT scan is an additional X-ray oriented medical imaging technology for examining internal organs [3]. The newer imaging technology is Magnetic Resonance Imaging (MRI). This is an advanced technology which exploits magnetic features and offers extensive information about inner structures [4]. The broader use of this technology in several medical application justifies its competency among previously used technologies. It provides 3-dimensional segments of each part with high contrast among tissues. However, the major hindrance in MRI applications is the high volume data which snag the manual analysis [5]. Consequently, MRI still has few drawbacks such as noise, low contrast among certain tissues, intensity in homogeneities, and partial volume effects in the segmentation task.

Image segmentation is vital in the computer vision especially in medical imaging [6]. More precisely, the image segmentation is the core part of the clinical image processing [7]–[9]. The segmentation of medical images is useful for accurate diagnoses of disease and for future treatment planning. It is used to locate tumors and other pathologies. Automated segmentation facilities computer-aided surgery. Furthermore, it can be used to measure tissue volume and to study their anatomical structure. While examining a disease, the image segmentation is a cardinal step in many syndromes [10]. For the finer investigation, the subdivision of the body (affected or non-affected) regions is imperative, which helps in decision making about the anomaly [10]. Various techniques have been developed for medical image segmentation. However, there is a sufficient space for improvement in the medical image segmentation. Majority of the existing segmentation methods are not fully automated and require some parameters setting to get accurate results. On the other hand, fully automated techniques have high computational complexity.

The present technologies in clinical image segmentations are specific according to their application, approach and body type. For instance, if we concern about brain segmentation, it has different technique from thorax segmentation technique [11]. Therefore, choosing an algorithm and technique for segmentation needs precise consideration [12]. There is no comprehensive algorithm for image segmentation in the field, consequently, every set of images has restrictions. D.J. Withy and Z.J. Koles provides a concise study of three different generations of image segmentation technologies [13]. Diagnostic imaging is an invaluable tool in medicine whereas Brain MRI is significant technology in present time. In medicine, the number of medical images is increasing. Acquisition of brain MRI may contain noise which can affect the accuracy of segmentation. However, a robust, efficient, and typologically precise segmentation of brain MRI is still a challenge. Brain MR Images consist of five tissues including background, skull, cerebrospinal fluid, gray and white matter. Each tissue has its own significance. In medicine, each tissue should be segmented accurately for diagnosis of diseases and to locate pathological condition. Therefore, a robust system is required to segment tissues of brain MRI images.

In the field of medical imaging, the segmentation has been used for alleviation of disparate tissues from each other, by withdrawing and categorizing characteristics. An exertion is pixel classification of an image into structural sections which is useful in separating muscles, bones, and blood containers. For instance, the objective of a brain study is to divide the full image into multiple sections based on colors. Therefore, brain image is segmented into grey scales to detect dissimilar tissues in brain i.e. cerebrospinal fluid, grey and white matter. Statistical and transform methods are applied to extract desired information from biomedical images [14]. Various techniques have been employed to segment magnetic resonance images which use different features extraction methods such as 2D-Discrete transform method [15]–[17], 2D-Continous wavelet transform method [18], the Gabor transform method [19] and gray level histogram [18] methods. A variety of classifiers are used to classify medical images [18], [20], [21]. However, there is lack of methodology for brain MRI segmentation. The major issue is that there is no adequate tool that can characterize different tissues of MRI images accurately. To this end, we propose a framework for brain MRI segmentation. We applied Gabor method to compute features of brain MR images and constructed a feature space. Then, four different classifiers are applied over feature space to classify the features into different segments. Due to different discriminative ability of the classifiers, their classification results in variant performance for brain MRI segmentation. Thus, we proposed a rule-based hybrid approach to improve the segmentation of brain tissues.

## II. METHODOLOGY

Gabor transform method is used to compute features of brain MR images. Then, different classifiers are used for pixel classification of these brain MR images.

### A. Gabor Transform Method (GTM)

Gabor function proposed by Dennis Gabor is frequently used in computer vision applications for feature extraction, especially in texture based image analysis like segmentation, classification and discrimination. The Gabor functions results in directional features for an image due to its capability of adjusting frequency properties, which reduces the noise in medical images [22].

The proposed workflow diagram is shown in Fig. 1. During offline training, Gabor features are computed using Gabor transform method and are fed to classifiers for training. Different classifiers are trained over Gabor features. After training these classifiers, test images are given to train models for segmentation. Rules are applied over segmented images to achieve improved segmentation.

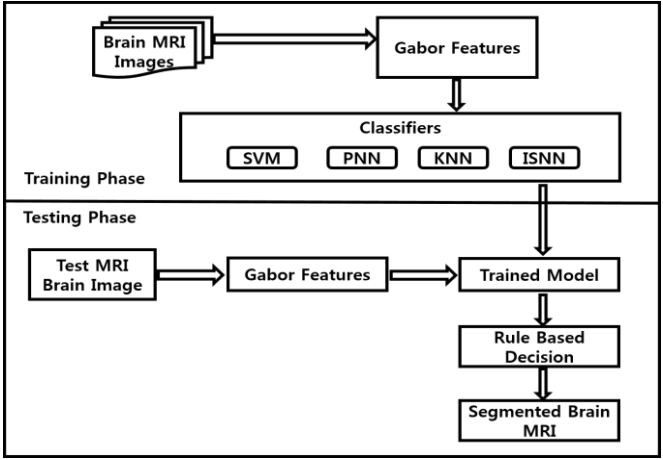

Figure 1: Image segmentation framework

### B. Probabilistic Neural Network (PNN)

A PNN is composed of four layers including input, pattern, summation and output layers as shown in Fig. 2. PNN algorithm is based on statistical algorithm known as kernel discriminant analysis. PNN is a feed forward network where inputs are processed in an organized manner to compute outputs. The PNN classifies the input pattern to class A based on the rule if

$$h_A C_A f_A(x) > h_B C_B f_B(x), \quad (1)$$

where $h_A$ is priori probability of occurrence of pattern in class A, $C_A$ denotes the cost of misclassification of the pattern of class A to pattern of class B and $f_A(x)$ represents the probability density function (PDF) for class A. The PDF for class A is calculated as:

$$f_A(x) = \frac{1}{(2\lambda)^{\frac{n}{2}}\sigma^n} \frac{1}{m} \sum_{i=1}^{m_A} exp \frac{(x-x_{Ai})^T(x-x_{Ai})}{2\sigma^2}, \quad (2)$$

where $n$ denotes the number of dimensions, $\sigma$ represents the smoothing parameter corresponding to standard deviation of the Gaussian distribution, $m_A$ represents the number of patterns of class A, and $x_{Ai}$ denotes the $i^{th}$ pattern of class A. A detailed discussion of PNN can be found at [20].

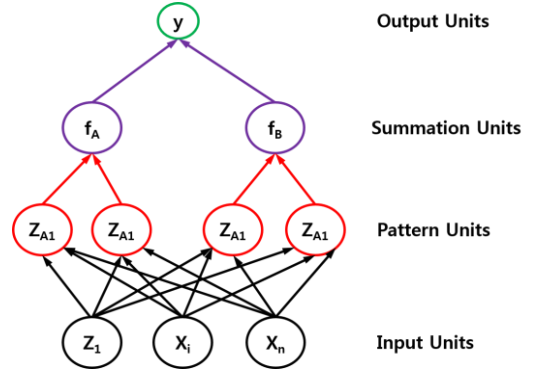

Figure 2: Architecture of PNN

*K-Nearest Neigbor (KNN)*

KNN is the simplest algorithm which is used for classification of objects. KNN algorithm classifies the objects based on the closest training features. It performs efficient nearest neighbor search and classifies the objects based on the majority voting among its neighbors [24].

### C. Incremental Supervised Neural Networks (ISNN)

ISNN is composed of two-layer network. Learning Vector Quantization (LVQ) network and ISNN has almost same structure. The only main difference between these two net is that whenever the input vector doesn't match with weight vector class during training of nets, then in LVQ weights are decreased while in ISNN input vector is added in first layer nodes. During training, the learning rate (μ) is kept constant [18].

### D. Support Vector Machine (SVM)

SVM is a supervised discriminative classifier but not a neural network which classifies the objects based on separating hyperplanes. SVM constructs a separating hyperplane among input features and performs classification based on those hyperplanes. SVM constructs hyperplanes such that it maximizes the distance among hyperplanes and separates the input data. For an error free binary classification, the hyperplanes have the maximum distance or the margin between the hyperplanes. The performance of the SVM is totally dependent upon training data and constructs separating hyperplanes and does not consider those data points that reside outside the margin [25].

Each classifier has different efficiency for tissue segmentation. Our objective is to develop a system so that we can get efficient segmentation result. We will compare efficiency of each classifier for each segmented tissue and will develop a rule based tissue segmentation method.

## III. EXPERIMENT AND RESULTS

Supervised classifiers are used to segment brain MR images using the Gabor transform method as feature extraction methods. The original image is shown in Fig. 3. Brain MRI is divided into five tissues as background, cerebrospinal fluid, gray-matter, white-matter and skull.

As mentioned earlier supervised learning is performed. There are 100 training points. These points are equally selected from each tissue for an image. So, we have training data with labels. A feature space has nine dimension thus, training data has size of 100 by 9 and training label size is 100 by 1. All the codes and simulations are run over inter(R) Core(TM) i3 CPU @ 2.93 GHz with MATLAB R2015b on Windows 7 Ultimate 64-bit.

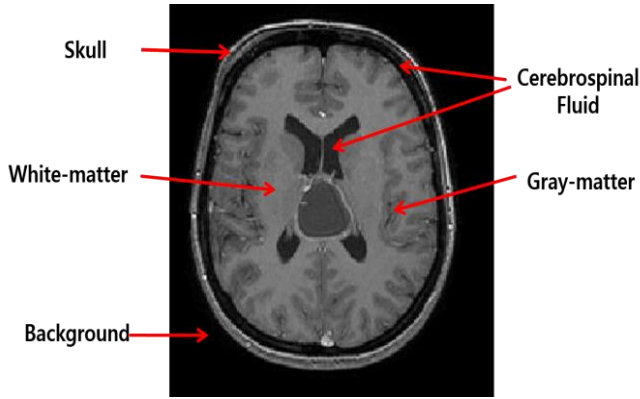

Figure 3: MRI brain image

The performance has been evaluated using Leave-One-Out Cross Validation (LOOCV) technique. In the field of machine learning, LOOCV is employed to validate the performance of the classification model and verify the prediction over testing data [23]. While using LOOCV, the classifier is trained on all training images, but one image is kept for testing. Test is performed on that excluded image. LOOCV is a useful approach to validate the performance of the trackers because it does not waste data. The main drawback of this approach is that it is computationally cost expensive because the training process is repeated multiple times [26].

Efficiency of the classifier is computed using f-measure techniques. In this technique, the true positive (TP), false negative (FN) and false positives (FP) are computed. TP are the pixel points that are present in segmented image and the original labeled image. FN are those pixel points that are present in the original labeled image but not in segmented image and FP are those points that are present in the segmented image but not in the original labeled image. Precision, recall and f-measure [27] are computed using formulas as given below:

$$\text{Precision} = \frac{TP}{TP+FP} \qquad (3)$$

$$\text{Recall} = \frac{TP}{TP+FN} \qquad (4)$$

$$F - \text{measure} = 2 \cdot \frac{Precision \cdot Recall}{Precision + Recall} \qquad (5)$$

The dataset consists of 11 patients MRI real images having size 128×128. The dataset is collected by Siemens MRI system in Abrar CT and MRI Centre, Peshawar road, Rawalpindi.

In LOOCV approach, training is performed over all the training images and only one image is kept for testing. In the 1st step, the classifiers are trained for all the 10 images except 1st image because this is kept for testing. Similarly, in 2nd step, 2nd image is kept for testing and so on. For each step f-measure is computed for each image. The groundtruth values for each image are provided by MRI image expert from dataset source. Fig. 4 shows the comparison of classifiers (SVM, PNN, KNN and ISNN) for all images using LOOCV approach. After averaging the performance of classifiers, SVM performs better than other classifiers as shown in Fig.5.

The output of ISNN, KNN, PNN and SVM classifiers for image segmentation are shown in Fig. 6. It can be observed that SVM has shown better performance than the other classifiers.

Performance of each classifier for each tissue is computed using F-measure. Results shows that segmentation of the background of MRI brain image for each classifier is 100%. But segmentation of remaining tissues varies according to each classifier. F-measure for SVM, PNN, KNN and ISNN for each tissue is computed as shown in Fig. 7.

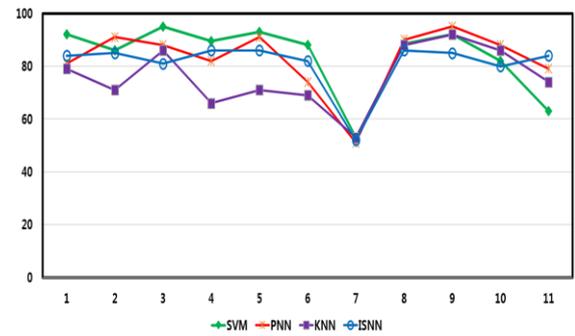

Figure 4: Comparison of classifiers using Gabor transform features

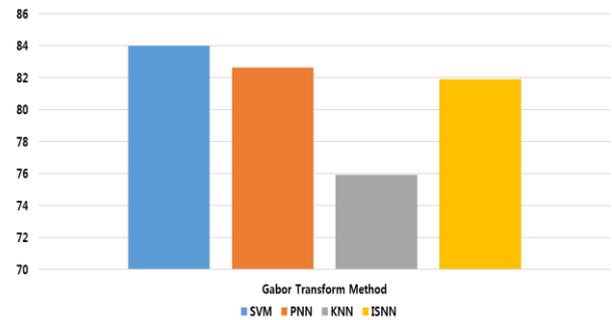

Figure 5: Comparison of classifiers

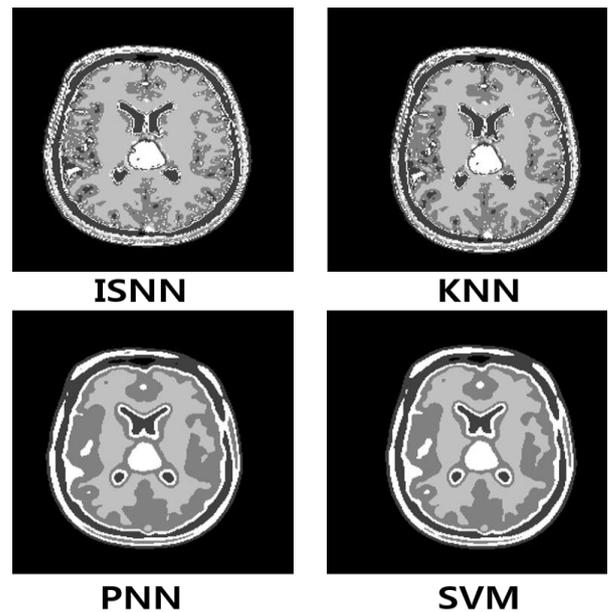

Figure 6: Segmented MRI brain images using ISNN, KNN, PNN, and SVM classifiers

Analyzing the results of different classifiers, a new rule-based hybrid approach is developed to classify different tissues as evident in Fig.8. Using this hybrid approach, to segment cerebrospinal fluid and skull tissues, SVM classifier should be used having efficiency of 87.72% and 91.82% for cerebrospinal fluid and skull tissues respectively, which is greater than efficiency of other classifiers for segmentation of these regions. Similarly, to segment gray matter, ISNN classifier has efficiency of 76.26% that is greater than any other classifier used for gray matter segmentation. For white matter segmentation, PNN and ISNN classifiers expressed same efficacy which is 85% greater than other classifiers efficiency.

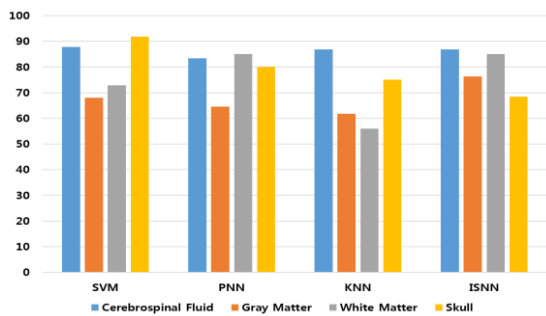

Figure 7: Comparison of classifiers with respect to their efficiency for brain MRI segmentation

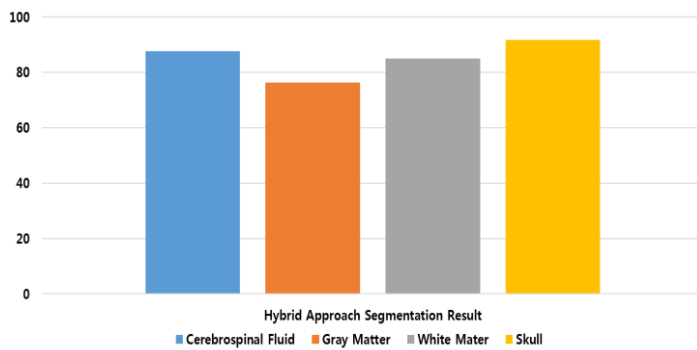

Figure 8: Hybrid approach segmentation results

IV. CONCLUSION

In this study, we proposed a framework for segmentation of brain MRI using Gabor transform method to build feature space and pixel classification of feature space is performed by four different classifiers i.e., SVM, ISNN, PNN and KNN. The evaluation of discriminative ability of these classifiers to perform segmentation of brain MR images indicated that all the classifiers have different ability of segmentation for different brain regions. So, we proposed a rule-based hybrid approach for better segmentation of brain MRI. By applying this approach, we conclude that SVM should be used for efficient segmentation of cerebrospinal fluid and skull tissue, ISNN should be employed for gray matter whereas, PNN and ISNN should be considered for better segmentation of white matter.

ACKNOWLEDGEMENT

This research was supported by the Basic Science Research Program through the National Foundation of Korea (NRF) funded by the Ministry of Education, Science and Technology (NRF-2016R1A2B1015101).